\documentclass[letterpaper]{article} 
\usepackage[]{aaai2026}  
\usepackage{times}  
\usepackage{helvet}  
\usepackage{courier}  
\usepackage[hyphens]{url}  
\usepackage{graphicx} 
\usepackage{natbib}  
\usepackage{caption} 
\usepackage{algorithm}
\usepackage{algorithmic}
\usepackage{amssymb,amsmath}
\usepackage{multirow}
\usepackage{subfigure}

\usepackage{booktabs}

\usepackage{color}

\usepackage{newfloat}
\usepackage{listings}
\DeclareCaptionStyle{ruled}{labelfont=normalfont,labelsep=colon,strut=off} 
\floatstyle{ruled}
\newfloat{listing}{tb}{lst}{}
\floatname{listing}{Listing}
\title{Reprogramming Vision Foundation Models for Spatio-Temporal Forecasting}
\author {
    Changlu Chen\textsuperscript{\rm 1},
    Yanbin Liu\textsuperscript{\rm 2},
    Chaoxi Niu\textsuperscript{\rm 3},
    Ling Chen\textsuperscript{\rm 3},
    Tianqing Zhu\textsuperscript{\rm 1} 
}
\affiliations {
    \textsuperscript{\rm 1}City University of Macau\\
    \textsuperscript{\rm 2}Auckland University of Technology\\
    \textsuperscript{\rm 3}University of Technology Sydney\\
    clchen@cityu.edu.mo, csyanbin@gmail.com, chaoxi.niu@student.uts.edu.au,\\
    ling.chen@uts.edu.au, tqzhu@cityu.edu.mo
}

\usepackage{bibentry}

\begin{document}

\maketitle

\begin{abstract}

Foundation models have achieved remarkable success in natural language processing and computer vision, demonstrating strong capabilities in modeling complex patterns. While recent efforts have explored adapting large language models (LLMs) for time-series forecasting, LLMs primarily capture one-dimensional sequential dependencies and struggle to model the richer spatio-temporal (ST) correlations essential for accurate ST forecasting.  
In this paper, we present \textbf{ST-VFM}, a novel framework that systematically reprograms Vision Foundation Models (VFMs) for general-purpose spatio-temporal forecasting. While VFMs offer powerful spatial priors, two key challenges arise when applying them to ST tasks: (1) the lack of inherent temporal modeling capacity and (2) the modality gap between visual and ST data. To address these, ST-VFM adopts a \emph{dual-branch architecture} that integrates raw ST inputs with auxiliary ST flow inputs, where the flow encodes lightweight temporal difference signals interpretable as dynamic spatial cues. 
To effectively process these dual-branch inputs, ST-VFM introduces two dedicated reprogramming stages. 
The \emph{pre-VFM reprogramming} stage applies a Temporal-Aware Token Adapter to embed temporal context and align both branches into VFM-compatible feature spaces. 
The \emph{post-VFM reprogramming} stage introduces a Bilateral Cross-Prompt Coordination module, enabling dynamic interaction between branches through prompt-based conditioning, thus enriching joint representation learning without modifying the frozen VFM backbone. 
Extensive experiments on ten spatio-temporal datasets show that ST-VFM outperforms state-of-the-art baselines, demonstrating effectiveness and robustness across VFM backbones (e.g., DINO, CLIP, DEIT) and ablation studies, establishing it as a strong general framework for spatio-temporal forecasting.

\end{abstract}

\section{Introduction}

Spatio-temporal forecasting aims to predict future dynamics by modeling the interactions between spatial locations and temporal evolution. It plays a crucial role across domains such as human mobility, traffic management, and urban computing~\cite{wang2020deep}. While deep learning models—such as CNNs, GCNs, RNNs, and Transformers~\cite{zhang2017deep,ref:AGCRN,ref:GMAN,zhou2023predicting}—have achieved notable success, they often require task-specific architectures and extensive retraining on exclusive datasets, limiting their scalability across diverse ST domains.

Foundation models have recently reshaped natural language processing (NLP) \cite{brown2020language,touvron2023llama} and computer vision (CV) \cite{dino,dino2}, demonstrating strong generalization from large-scale pretraining. 
Motivated by the sequential nature shared between language and time-series data, several studies have explored adapting large language models (LLMs) for time-series forecasting~\cite{ref:time-llm,cao2023tempo,zhou2023one} and even early-stage ST forecasting~\cite{liu2024spatial,huang2024std}. 
However, these adaptations are fundamentally limited: LLMs are designed to capture one-dimensional sequential dependencies and cannot naturally handle the multi-dimensional spatio-temporal correlations that characterize ST data. 

In contrast, \textbf{Vision Foundation Models (VFMs)}, particularly Vision Transformers (ViTs), pretrained on massive image datasets, offer a promising alternative due to their strong spatial modeling capabilities. Notably, spatio-temporal grids $X \in \mathbb{R}^{H \times W \times C\times T}$ structurally resemble visual grids $\mathbb{R}^{H \times W \times C}$ at each time step $t$, suggesting that VFMs can be leveraged for ST forecasting. 
However, two main challenges arise when applying VFMs to ST tasks:
(1) the lack of native temporal modeling capacity, and (2) the modality gap between visual and ST data. 

To address these challenges, we propose \textbf{ST-VFM}, a novel framework that systematically reprograms VFMs for general-purpose spatio-temporal forecasting. ST-VFM adopts a \emph{dual-branch architecture} that integrates (i) raw ST inputs capturing spatial patterns and (ii) auxiliary ST flow inputs encoding lightweight temporal difference signals, allowing VFMs to perceive temporal dynamics as spatial cues.

To enable VFMs to effectively process the dual-branch spatio-temporal inputs, we introduce two dedicated reprogramming stages. 
The \emph{pre-VFM reprogramming} stage employs a Temporal-Aware Token Adapter that projects both the raw ST inputs and auxiliary ST flow signals into token representations compatible with the VFM’s spatial encoder, embedding temporal context in a form that VFM can interpret. 
Subsequently, the \emph{post-VFM reprogramming} stage integrates a Bilateral Cross-Prompt Coordination mechanism, allowing the two branches to exchange complementary spatial and temporal cues through prompt-based interaction, thereby enriching joint representation learning without modifying the frozen VFM backbone.

We conduct comprehensive experiments on ten diverse ST datasets across domains including traffic, mobility, crowd flow, and cellular usage. ST-VFM consistently outperforms prior state-of-the-art methods, demonstrating the effectiveness of reprogrammed VFMs as a flexible and general backbone for spatio-temporal forecasting. We also validate the framework across multiple VFM backbones (e.g., DINO, CLIP, DEIT) and provide extensive ablation studies.

Our main contributions are summarized as follows: 
\begin{itemize}
    \item We propose \textbf{ST-VFM}, the first framework to systematically reprogram Vision Foundation Models for general-purpose spatio-temporal forecasting across diverse domains.
    
    \item We design a novel \emph{dual-branch architecture} that integrates raw spatio-temporal inputs with auxiliary ST flow signals, enabling VFMs to jointly capture spatial patterns and lightweight temporal dynamics.
    
    \item We develop an effective two-stage reprogramming strategy, combining \emph{Temporal-Aware Token Adapter} and \emph{Bilateral Cross-Prompt Coordination}, to adapt non-image inputs to the spatial inductive biases of VFMs while keeping the backbone frozen. 
    
    \item Extensive experiments are conducted on ten benchmarks spanning traffic, mobility, demand, and communication datasets, demonstrating state-of-the-art performance, cross-backbone flexibility (e.g., DINO, CLIP, DEIT), and robust improvements validated through ablation studies.

\end{itemize}

\section{Related Work}
\subsection{Spatio-Temporal Forecasting}

Spatio-temporal forecasting aims to predict future dynamics by modeling the interdependencies between spatial locations and temporal sequences in historical data, with applications across domains such as traffic, human mobility, and communication networks~\cite{wang2020deep}. Early approaches primarily relied on statistical models, but they struggled to capture the nonlinear and high-dimensional nature of ST data.

In recent years, deep learning has become the dominant paradigm. Recurrent neural networks (RNNs)~\cite{chung2014empirical} have been widely used to model temporal dependencies, while graph convolutional networks (GCNs)~\cite{yu2017spatio,li2017diffusion} and convolutional neural networks (CNNs)~\cite{ref:GSNet} have been integrated to capture spatial correlations. More recently, Transformer-based architectures~\cite{ref:GMAN,jiang2023pdformer} have shown superior capability in jointly modeling spatial and temporal patterns through self-attention mechanisms.

Beyond architecture design, advanced techniques like self-supervised learning~\cite{ji2023spatio,zhang2023mask} and transfer learning~\cite{jin2022selective,lu2022spatio} have been explored to improve representation learning by leveraging auxiliary tasks or cross-domain knowledge.  

However, most of these approaches are tailored to specific domains and require full retraining on targeted datasets, limiting their generalization to diverse ST forecasting scenarios. In contrast, our approach reprograms pretrained VFMs to capture both spatial and temporal correlations, enabling effective and unified ST forecasting across domains.

\subsection{Foundation Models for Spatio-Temporal Forecasting}
With the rise of large language models and their powerful reasoning and generalization capabilities, recent work has explored applying LLMs to spatio-temporal forecasting. For example, ST-LLM~\cite{liu2024spatial} treats spatial locations as language tokens and introduces an embedding layer to align spatio-temporal representations with the LLM tokenization scheme. STG-LLM~\cite{liu2024can} views each graph node as a token, enabling LLMs to represent both spatial and temporal correlations through graph semantics. STD-PLM~\cite{huang2024std} designs dedicated spatial and temporal tokenizers to convert ST data into sequential inputs suitable for pre-trained language models.

Beyond language models, OpenCity~\cite{li2024opencity} pre-trains a Transformer-Graph architecture on large-scale traffic datasets, learning transferable ST representations across heterogeneous domains. UrbanGPT~\cite{li2024urbangpt} integrates an ST dependency encoder with instruction tuning to better align spatial and temporal dependencies. UrbanCLIP~\cite{yan2024urbanclip} aligns language and vision representations by pairing urban imagery with generated textual descriptions using contrastive learning.

Notably, UniST~\cite{ref:UniST} represents a recent foundation model specifically pre-trained on diverse spatio-temporal datasets and fine-tuned for target ST forecasting tasks, demonstrating strong generalization across domains.

While these approaches improve generalization for ST forecasting, they predominantly rely on language models or require domain-specific pre-training on ST data. To our knowledge, no prior work has explored leveraging VFMs for simultaneous spatio-temporal representation learning. In contrast, our approach explicitly adapts VFMs to capture both spatial and temporal correlations, offering a unified and effective solution for general-purpose ST forecasting.

\begin{figure*}[t]
\centering
\includegraphics[width=1.0\textwidth]{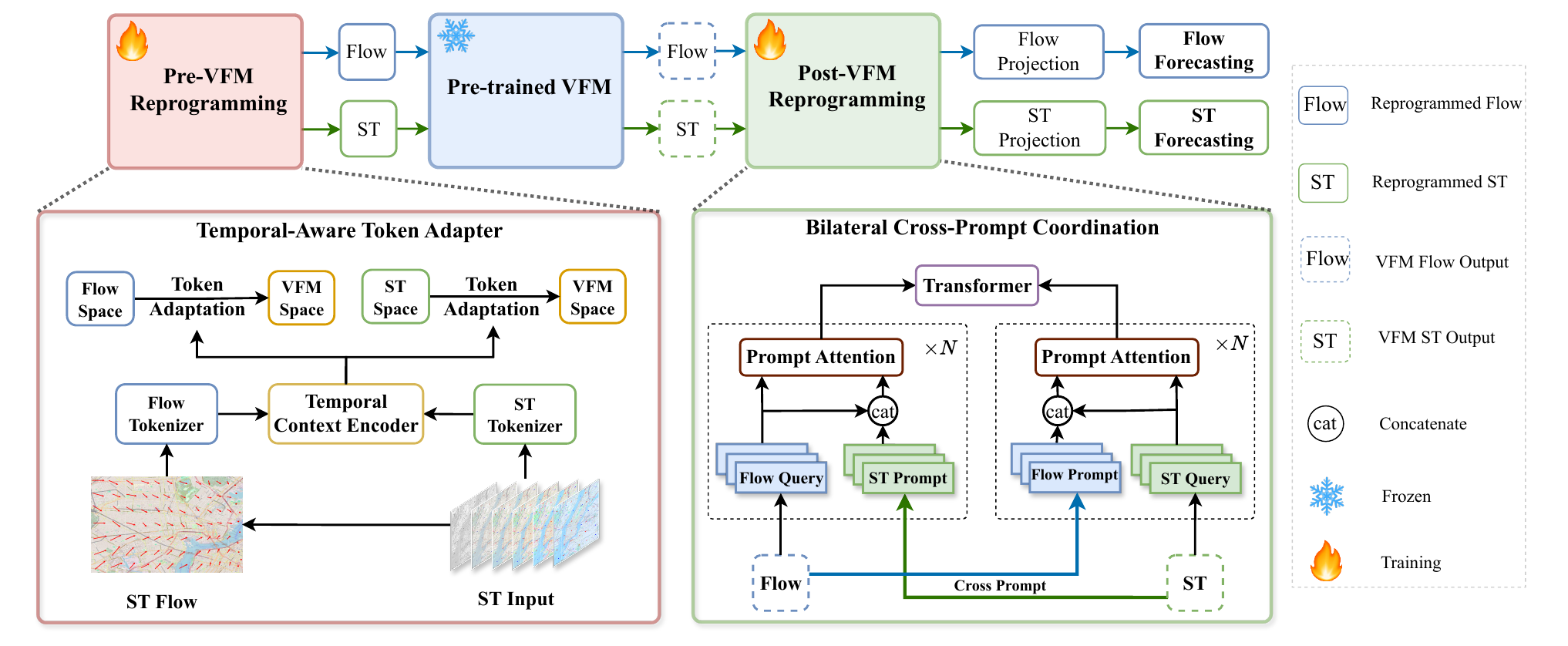}  
\caption{\label{fig:framework}
\textbf{The overall framework of ST-VFM}. 
ST-VFM takes two input sources: \emph{spatio-temporal input} and \emph{ST flow}. 
In the Pre-VFM Reprogramming stage, a \emph{Temporal-Aware Token Adapter} projects both inputs into a VFM-compatible token space, embedding temporal context while aligning modalities. 
The frozen VFM then encodes these adapted ST and flow tokens. 
In the Post-VFM Reprogramming stage, a \emph{Bilateral Cross-Prompt Coordination} module enables bidirectional interaction between the VFM outputs from both branches: each output is conditioned on contextual prompts derived from the other branch via prompt attention blocks. 
The output features support ST forecasting, with flow forecasting serving as an auxiliary training objective. 
}
\end{figure*}

\section{Methodology}
\subsection{Preliminary}

\paragraph{Spatio-Temporal Data}
We consider spatio-temporal data collected from a target region partitioned into a uniform $H \times W$ grid based on geographic coordinates. Each grid cell aggregates domain-specific features (e.g., traffic volume, bike usage, cellular activity), forming a 4D tensor $\mathcal{X} \in \mathbb{R}^{C \times H \times W \times T}$, where $T$ is the number of time steps, $C$ is the feature dimension, and $H$, $W$ denote the spatial dimensions. This structure captures evolving spatial and temporal dependencies across the region.

\paragraph{Spatio-Temporal Forecasting}
Given historical observations $\mathcal{X}^P = (X_1, \ldots, X_P) \in \mathbb{R}^{C \times H \times W \times P}$ over $P$ time steps, the objective is to predict future values over the next $Q$ steps, denoted as $\mathcal{X}^Q = (X_{P+1}, \ldots, X_{P+Q}) \in \mathbb{R}^{C \times H \times W \times Q}$. This forecasting task is modeled by a learnable model $\mathcal{F}_\theta$ parameterized by $\theta$:
\begin{equation}
\mathcal{X}^Q = \mathcal{F}_\theta(\mathcal{X}^P)\,.
\end{equation}
This setting encompasses diverse urban prediction tasks, such as traffic speed forecasting, crowd flow estimation, cellular usage modeling, and mobility demand prediction.

\subsection{Motivation and Framework Overview}

Existing spatio-temporal forecasting models often rely on architectures tailored to specific domains~\cite{li2017diffusion,ref:GMAN,jiang2023pdformer}, limiting their generalization to diverse tasks. In contrast, VFMs~\cite{dino,ref:CLIP} offer transferable spatial representations learned from large-scale natural image corpora, yet remain underexplored for ST prediction.

ST data exhibits structural similarities to visual data: each frame $X_t \in \mathbb{R}^{C \times H \times W}$ resembles an image, and a sequence of such frames forms a video-like tensor. This motivates leveraging VFMs as powerful spatial encoders for ST forecasting. However, two key challenges arise: \textbf{(1)} VFMs are not designed for temporal reasoning or for aligning with ST semantics (e.g., traffic flow or demand volume). \textbf{(2)} Raw ST data does not conform to typical VFM input assumptions, such as RGB channels or spatial patching schemes, making direct application infeasible. 

To address these limitations, we propose \textbf{ST-VFM}, a dual-branch framework that adapts VFMs for spatio-temporal forecasting. 
The dual branches process \textbf{(i)} the raw ST input and \textbf{(ii)} the derived ST Flow, which captures frame-wise dynamics. 
These inputs are reprogrammed through:
\begin{itemize}
    \item Pre-VFM Reprogramming: A \textbf{\emph{Temporal-Aware Token Adapter}} projects both the ST input and the ST Flow into a VFM-compatible token space, aligning non-image inputs with the spatial inductive biases of VFMs while embedding temporal context through flow features.

    \item Post-VFM Reprogramming: A \textbf{\emph{Bilateral Cross-Prompt Coordination}} module enables bidirectional interaction between the VFM outputs from both branches, where each output is conditioned on contextual prompts from the other branch via prompt attention, enhancing the joint modeling of spatial and temporal representations.
\end{itemize}

\subsection{Pre-VFM Reprogramming}

\subsubsection{Two-Branch Input Design with ST Flow}
To address the lack of temporal modeling in image-pretrained VFMs, one might consider video foundation models (Video FMs). However, existing Video FMs often underperform compared to large-scale image-based VFMs,\footnote{See Table~\ref{tab:video} for comparison.} due to smaller pretraining datasets and weaker separation of spatial and temporal cues.
 
Instead, we draw on video processing practices, where two-stream architectures~\cite{Two-stream,Flow} combine RGB frames with optical flow to enhance temporal modeling. Following this principle, we design a two-branch input: the \textbf{raw ST branch}, providing spatial context, and the \textbf{ST Flow branch}, using simplified temporal difference maps as a lightweight motion proxy. This enables image-pretrained VFMs to integrate temporal signals efficiently without architectural modification. 

We derive the ST Flow as a lightweight approximation of temporal dynamics, following optical flow principles. Given $X \in \mathbb{R}^{H \times W \times T}$, optical flow estimates the per-pixel displacement field $F(h, w, t) = (u(h, w, t), v(h, w, t))$ under the brightness constancy assumption: 
\begin{equation}
    X(h, w, t) = X(h + u(h, w, t),\, w + v(h, w, t),\, t + 1),
\end{equation}
where $u(h, w, t)$ and $v(h, w, t)$ are the horizontal and vertical displacements between time steps $t$ and $t + 1$. Applying first-order Taylor expansion yields the flow constraint:
\begin{equation}
    X_h \cdot u + X_w \cdot v + X_t = 0,
\end{equation}
where $X_h$, $X_w$, $X_t$ denote partial derivatives with respect to $h$, $w$, and $t$. Estimating $u$ and $v$ typically requires smoothness priors or local patch optimization~\cite{lucas1981iterative}, which is computationally intensive and sensitive to noise. 
To avoid these complexities, we focus solely on the temporal derivative term:
\begin{equation}
    \Delta_t X(h, w, t) = X(h, w, t + 1) - X(h, w, t),
\end{equation}
which serves as a first-order approximation of local temporal variation. While it omits explicit displacement direction, it effectively captures motion magnitude and dynamic change patterns. We incorporate the resulting ST Flow as an auxiliary input branch, reprogrammed into the VFM token space to complement the raw spatial inputs with temporal cues.

\subsubsection{Temporal-Aware Token Adapter} 
Adapting ST data to VFMs requires addressing the modality gap between spatio-temporal inputs $C \times H \times W \times T$ and natural image inputs $C \times H \times W$. While the spatial dimensions share a grid structure, the additional temporal axis $T$ carries dynamic patterns that cannot be captured by standard ViT-style tokenization. To handle this, we introduce the Temporal-Aware Token Adapter with three components: 
 
\paragraph{(1) \emph{ Flow/ST Tokenizer}}  
Following practices in ViT \cite{ref:ViT} and recent spatio-temporal models~\cite{ref:UniST}, the Flow/ST Tokenizer $\mathcal{E}_{\mathrm{token}}$ partitions the input tensor $X \in \mathbb{R}^{C \times H \times W \times T}$ into non-overlapping spatial patches of size $P_s \times P_s$ along the $H, W$ dimensions at each time step $t$. Each patch is flattened into a vector of dimension $P_s^2 \cdot C$, producing $N_s$ spatial tokens per frame and forming a temporal sequence $\{ X_t \}_{t=1}^T$. This patch-based tokenization aligns with the inductive biases of ViT-style VFMs and reduces computational cost for long-range spatial modeling. However, since it does not capture temporal dependencies across frames, we introduce a dedicated Temporal Context Encoder below to model relationships along the $T$ dimension.

\paragraph{(2) \emph{Temporal Context Encoder}}  
Given the output from the Flow/ST Tokenizer, represented as $X \in \mathbb{R}^{N_s \times (P_s^2 \cdot C) \times T}$, we first apply a linear projection:
\begin{equation}
    Z_{\mathrm{proj}} = X \cdot W_{\mathrm{proj}}, \quad W_{\mathrm{proj}} \in \mathbb{R}^{(P_s^2 \cdot C) \times D_T},
\end{equation}
which maps the patch tokens into a temporal embedding space $Z_{\mathrm{proj}} \in \mathbb{R}^{N_s \times D_T \times T}$.  
We then apply a lightweight temporal Transformer across the $T$ dimension (shared across all $N_s$ spatial locations) in a batched fashion:
\begin{equation}
    Z_{\mathrm{temp}} = \mathrm{TemporalTrans}\big(Z_{\mathrm{proj}}\big),
\end{equation}
where $\mathrm{TemporalTrans}$ denotes a temporal self-attention block operating over the last dimension.  
This produces temporally contextualized representations $Z_{\mathrm{temp}} \in \mathbb{R}^{N_s \times D_T \times T}$, ensuring that temporal dynamics are explicitly embedded before adaptation to the VFM input space. Together with the ST Flow branch, this temporal modeling augments the VFM's inherent inability to capture temporal dependencies, providing complementary cues for effective spatio-temporal forecasting.

\paragraph{(3) \emph{Token Adaptation}} 
The final stage, Token Adaptation $\mathcal{E}_{\mathrm{adapt}}$, aligns the temporally enriched spatio-temporal features with the input space of the pre-trained VFM. Specifically, we apply a linear projection:
\begin{equation}
    Z_{\mathrm{vfm}} = Z_{\mathrm{temp}} \cdot W_{\mathrm{adapt}}, \quad W_{\mathrm{adapt}} \in \mathbb{R}^{D_T \times D_{\mathrm{VFM}}},
\end{equation}
where $Z_{\mathrm{temp}} \in \mathbb{R}^{N_s \times D_T \times T}$ is reshaped into $Z_{\mathrm{temp}} \in \mathbb{R}^{(N_s \cdot T) \times D_{\mathrm{T}}}$ by merging the temporal dimension into the batch axis. 
This produces token representations that are fully compatible with the frozen VFM backbone, enabling it to process spatio-temporal inputs using its pretrained visual knowledge from large-scale datasets.

\paragraph{\emph{Learnable Positional Embedding}}  
Before inputting the adapted tokens into the VFM, it is crucial to incorporate spatial positional information. While pre-trained VFMs include positional embeddings learned on natural images (e.g., $224 \times 224$), prior approaches often interpolate these embeddings to match downstream tasks. However, as shown in Table~\ref{tab:datasets}, spatio-temporal datasets typically have much smaller grids (e.g., $16 \times 20$), making naive interpolation suboptimal.  

To address this, we introduce a learnable positional embedding $E^{\mathrm{pos}} \in \mathbb{R}^{N_s \times D_{\mathrm{VFM}}}$, \emph{initialized independently and broadcast across all $T$ frames}. The positional embedding is added to the adapted token representations as:
\begin{equation}
    Z_{\mathrm{vfm}} = Z_{\mathrm{vfm}} + E^{\mathrm{pos}},
\end{equation}
where $E^{\mathrm{pos}}$ is repeated along the temporal dimension to match the shape of $Z_{\mathrm{vfm}}$.  
This enables the model to tailor its spatial priors to the unique characteristics of spatio-temporal data, improving downstream spatial pattern recognition.

\subsection{Post-VFM Reprogramming}

Following the Pre-VFM Reprogramming stage, we introduce a \emph{Bilateral Cross-Prompt Coordination} module to enable mutual enhancement between the ST and Flow branches. Unlike prior prompt tuning approaches that use learned or randomly initialized prompts, we construct \textbf{data-driven cross-prompts}, dynamically extracted from the VFM outputs of each branch. This mechanism allows each branch to inject complementary information into the other, improving joint spatio-temporal modeling.

Specifically, we define the cross-prompt tokens as:
\begin{equation}
    P^{\mathrm{ST}} = \{ p_j^{\mathrm{ST}} \}_{j=1}^{L_p}, \quad P^{\mathrm{Flow}} = \{ p_j^{\mathrm{Flow}} \}_{j=1}^{L_p},
\end{equation}
where $p_j^{\mathrm{ST}} \in \mathbb{R}^{D}$ and $p_j^{\mathrm{Flow}} \in \mathbb{R}^{D}$ are prompt vectors extracted from the ST and Flow branch outputs, respectively.

For the ST branch, we compute the attention weights conditioned on Flow-derived cross-prompts as:
\begin{equation}
    A_{i,j}^{\mathrm{ST}} = \frac{ e^{ Q_i^{\mathrm{ST}} \cdot [ K_j^{\mathrm{ST}}, P_j^{\mathrm{Flow}} ]^T } }{ 
    \sum_{k=1}^m e^{ Q_i^{\mathrm{ST}} \cdot (K_k^{\mathrm{ST}})^T } + \sum_{k=1}^{L_p} e^{ Q_i^{\mathrm{ST}} \cdot (P_k^{\mathrm{Flow}})^T }
    },
\end{equation}
where $Q_i^{\mathrm{ST}}$ is the $i$-th query, $K_j^{\mathrm{ST}}$ is the $j$-th key, and $P_j^{\mathrm{Flow}}$ is the $j$-th cross-prompt key; $m$ and $L_p$ are the numbers of regular and prompt tokens, respectively.  
The Flow branch applies an analogous formulation, conditioning on $P^{\mathrm{ST}}$ to enhance its temporal representation.

\paragraph{Rationale and interpretability.}  
This bilateral design is tightly aligned with the underlying structure of the spatio-temporal data (i.e., two-branch inputs):
\begin{itemize}
    \item $\Delta_t X$ (Flow) highlights temporal differences across frames, directing the ST branch to attend to dynamic and evolving patterns. 
    \item $X_t$ (ST) provides stable spatial context, helping the Flow branch anchor displacement estimates to meaningful structures.
\end{itemize}

For example, the Flow-derived prompts inform how $X_t$ evolves from $X_{t-1}$, while the ST-derived prompts highlight which spatial regions persist over time. By cross-referencing these complementary views, the model effectively disentangles motion and appearance, leveraging their interplay for improved forecasting.

Importantly, because the cross-prompts are dynamically computed from the data, they provide instance-specific, interpretable conditioning. This enables adaptive and flexible coordination between branches, extending the representational capacity of the frozen VFM without modifying its backbone architecture.

\subsection{Training Objective}

\paragraph{Decoder}  
After the dual-branch encoding and cross-prompt coordination, the model produces spatio-temporal feature representations from both the ST and Flow branches. To forecast future sequences, we apply a temporal Transformer decoder over the historical features, separately for each branch, to capture temporal correlations and generate predictive representations.

\paragraph{Auxiliary Flow Forecasting}
Beyond the main spatio-temporal forecasting task, we introduce an auxiliary flow forecasting objective. While the ST branch predicts the future spatio-temporal states, the Flow branch is trained to predict the future temporal difference (flow) maps, which serve as a lightweight approximation of first-order temporal dynamics. This auxiliary task provides additional supervision, encouraging the model to explicitly disentangle static spatial context from dynamic temporal change. Importantly, it acts as a regularizer, guiding the dual-branch architecture to better leverage complementary signals and reducing the risk of overfitting to either modality.

\paragraph{Loss Function}  
The model is jointly trained using Mean Squared Error (MSE) losses on both tasks:
\begin{equation}
    \mathcal{L} = \mathcal{L}_{\mathrm{ST}} + \lambda \mathcal{L}_{\mathrm{Flow}},
\end{equation}
where $\lambda$ controls the contribution of the auxiliary flow loss. This multi-task formulation encourages mutual reinforcement between the two branches, improving the robustness and accuracy of spatio-temporal forecasting.

\section{Experiments}

\begin{table}[htbp]
\centering
\setlength{\tabcolsep}{1.5pt}
\resizebox{0.5\textwidth}{!}{
\begin{tabular}{lcccccc}
\hline
Dataset & Domain & City & Temporal Span & Interval & Grid \\
\hline
Crowd        & Crowd Flow       & Nanjing    & 20.11.11--21.05.31 & 0.5H   & $16 \times 20$ \\
Cellular     & Cellular Usage   & Nanjing    & 20.11.11--21.05.31 & 0.5H   & $16 \times 20$ \\
TDrive       & Taxi Trajectory  & NYC        & 15.02.01--16.06.02 & 1H     & $32 \times 32$ \\
BikeNYC-1    & Bike Usage       & NYC        & 16.08.01--16.09.29 & 1H     & $16 \times 8$  \\
BikeNYC-2    & Bike Usage       & NYC        & 16.07.01--16.08.29 & 0.5H   & $10 \times 20$ \\
TaxiNYC-1    & Taxi OD          & NYC        & 16.01.01--16.02.29 & 0.5H   & $16 \times 12$ \\
TaxiNYC-2    & Taxi OD          & NYC        & 15.01.01--15.03.01 & 0.5H   & $20 \times 10$ \\
TrafficZZ    & Traffic Speed    & Zhengzhou  & 22.03.05--22.04.05 & 5 min  & $26 \times 26$ \\
TrafficTJ    & Traffic Speed    & Tianjin    & 22.03.05--22.04.05 & 5 min  & $24 \times 30$ \\
TrafficSH    & Traffic Speed    & Shanghai   & 22.01.27--22.02.27 & 5 min  & $28 \times 32$ \\
\hline
\end{tabular}}
\caption{Statistics of the ten spatio-temporal datasets used in our experiments.}
\label{tab:datasets}
\end{table}

\begin{table*}[!htbp]
\centering
\setlength{\tabcolsep}{2.2pt}
\resizebox{\textwidth}{!}{%
\begin{tabular}{c|llllllllllllllllllll}
\hline
\multirow{2}{*}{Methods} & \multicolumn{2}{c}{Crowd} & \multicolumn{2}{c}{Cellular} & \multicolumn{2}{c}{TDrive}      & \multicolumn{2}{c}{BikeNYC} & \multicolumn{2}{c}{BikeNYC2} & \multicolumn{2}{c}{TaxiNYC} & \multicolumn{2}{c}{TaxiNYC2} & \multicolumn{2}{c}{TrafficZZ} & \multicolumn{2}{c}{TrafficTJ} & \multicolumn{2}{c}{TrafficSH}\\ \cline{2-21} 
& RMSE         & MAE        & RMSE          & MAE          & RMSE & MAE & RMSE          & MAE         & RMSE          & MAE          & RMSE          & MAE         & RMSE          & MAE          & RMSE           & MAE          & RMSE           & MAE          & RMSE           & MAE          \\ \hline
HA & 17.80 & 6.79 &72.94 &27.57 &150.2 &74.5 &11.41 &3.43 &15.68 &7.17 &57.07 &18.57 &52.84 &15.74 &1.47 &0.857 &1.61 &0.824 &1.24 &0.771 \\
ARIMA &21.87 &10.23 &81.31 &40.22 & 211.3 &108.5 &12.37 &3.86 &25.01 &13.63 &55.39 &20.94 &62.9 &29.56 &1.78 &0.998 &2.02 &1.59 &1.17 &0.769\\
STResNet &5.355 &3.382 &24.30 &14.32 &220.1 &117.4 &8.20 &4.98 &7.18 &3.94 &29.45 &17.96 &22.16 &12.06 &1.03 &0.693 &1.12 &0.714 &1.00 &0.723\\
ACFM &4.17 &2.34 &22.79 &12.00 &98.1 &51.9 &3.93 &1.67 &5.99 &3.094 &23.35 &11.54 &14.48 &6.39 &0.839 &0.526&0.959 &0.574  &0.833 &0.566\\
STID &3.85 &1.63 &18.77 &8.24 &47.4 &23.3 &4.06 &1.54 &5.70 &2.711 &17.75 &7.03 &17.37 &7.35 &0.838 &0.502&0.976 &0.549 &0.742 &0.469\\
STNorm &4.44 &2.09 &19.77 &8.19 &54.3 &47.9 &4.45 &1.66  &6.47 &3.03 &21.26 &8.14 &19.02 &7.17&0.885 &0.538&0.973 &0.533 &0.871 &0.579\\
STGSP &7.93 &4.56 &39.99 &21.40 &94.6 &47.8 &5.00 &1.69 &14.20 &7.38 &28.13 &10.29 &29.10 &10.14 &0.831 &0.505&0.989 &0.572 &1.02 &0.749\\
MC-STL &4.75 &2.39 &21.22 &10.26 &54.2 &28.1 &4.08 &2.05 &6.26 &3.40 &18.44 &9.51 &16.78 &8.50 &1.14 &0.81&1.22 &0.856 &1.00 &0.720\\
MAU &4.94 &2.35 &39.09 &18.73 &5.22 &2.06 &1.28 &22.1 &6.12 &2.95 &28.70 &11.23 &19.38 &7.27 &1.42 &0.934&0.988 &0.549 &1.37 &0.991\\
PredRNN &5.13 &2.36 &24.15 &10.44 &54.9 &25.2 &5.00 &1.74 &6.47 &3.08 &16.53 &5.80 &19.89 &7.23 &0.853 &0.508&0.971 &0.53 &0.748 &0.469\\
MIM &5.66 &2.27 &21.38 &9.37 &51.4 &22.7 &4.40 &1.62 &6.36 &2.89 &18.83 &6.866 &18.02 &6.56 &2.05 &1.56&3.44 &2.51 &0.760 &0.505\\
SimVP &3.91 &1.96 &16.48 &8.23 &46.8 &22.9 &4.11 &1.67 &5.96 &2.92 &16.63 &7.51 &15.10 &6.54 &0.838 &0.526&1.00 &0.597 &0.814 &0.569\\
TAU &4.09 &2.11 &17.94 &8.91 &51.6 &28.1 &4.30 &1.83 &5.98 &2.89 &16.91 &6.85 
&15.35 &6.80 &0.839 &0.527 &1.01 &0.606 &0.820 &0.557\\
PatchTST &10.25 &3.62 &43.40 &15.74 &106.40 &51.30 &5.27 &1.65 &12.33 &5.30 &41.34 &13.10 &37.76 &11.13 &1.31 &0.742 &1.44 &0.722 &1.10 &0.663 \\
iTransformer &9.40 &3.40 &37.01 &13.93 &86.3 &42.6 &7.74 &2.53 &9.86 &4.50 & 36.73 & 13.11 &33.03 &11.22 &1.19 &0.696 &1.26 &0.675 &1.04 &0.655\\
ST-LLM & 3.17 & 1.26 & 15.04 & 6.01 &41.50 &19.55 & 7.65 & 4.34 & 7.75 & 4.58 & 21.25 & 9.64 & 19.02 & 8.65 & 0.907 & 0.576 &1.07 &0.65 &0.710 & 0.457 \\
UniST & 3.00  & 1.38 &   14.29 & 6.59 & 44.97 & 19.67 & 3.50 & 1.27 & 5.50 & 2.56 & 15.32 & 5.65 & 12.71 & 4.82 & 0.832 & 0.482 &  0.958 & 0.510 & 0.665 & 0.405 \\ \hline
ST-VFM &\textbf{2.51} &\textbf{1.06} &\textbf{12.31} & \textbf{5.57} &\textbf{39.54} & \textbf{16.94} &\textbf{3.38}  &\textbf{1.20} & \textbf{5.42} &\textbf{2.46}  &\textbf{13.79} &\textbf{4.95} &\textbf{11.85} & \textbf{4.39} & \textbf{0.800} & \textbf{0.457} & \textbf{0.909} & \textbf{0.484} & \textbf{0.645} &\textbf{0.392} \\ \hline
\end{tabular}
}
\caption{Comparison with state-of-the-art methods across diverse datasets, evaluated by MAE and RMSE (lower values indicate better performance). Our method outperforms all baselines on all datasets and metrics.}
\label{tab:result}
\end{table*}

\subsection{Experimental Setup}
\paragraph{Dataset \& Metrics}
We conduct experiments on ten diverse spatio-temporal datasets spanning multiple cities and application domains, following the evaluation protocol from~\cite{ref:UniST}. These datasets cover a wide range of tasks, including traffic speed forecasting, crowd flow prediction, taxi demand estimation, and cellular usage forecasting. They differ in input channel dimensions, temporal sampling intervals, spatial grid resolutions, temporal spans, and dataset sizes, as summarized in Table~\ref{tab:datasets}.

For fair comparison with prior work, the results on the Crowd and Cellular datasets are scaled by $10^{3}$ following~\cite{ref:UniST}. We divide each dataset into training, validation, and test splits using a standard $70\%$/$15\%$/$15\%$ ratio. We evaluate forecasting performance using two widely adopted metrics: Mean Absolute Error (MAE) and Root Mean Squared Error (RMSE). 

\paragraph{Baselines}
We compare our method against a comprehensive set of state-of-the-art spatio-temporal forecasting baselines, organized into five categories:
(1) \textbf{\emph{Conventional models}}: History Average (HA) and ARIMA;
(2) \textbf{\emph{Deep learning-based spatio-temporal models}}: STResNet~\cite{zhang2017deep}, ACFM~\cite{liu2018attentive}, MC-STL~\cite{zhang2023mask}, STGSP~\cite{zhao2022st}, STNorm~\cite{deng2021st}, and STID~\cite{shao2022spatial};
(3) \textbf{\emph{Video prediction models}}: PredRNN~\cite{wang2017predrnn}, MAU~\cite{chang2021mau}, MIM~\cite{wang2019memory}, SimVP~\cite{gao2022simvp}, and TAU~\cite{tan2023temporal};
(4) \textbf{\emph{Multivariate time series forecasting models}}: PatchTST~\cite{nie2022time} and iTransformer~\cite{liu2023itransformer};
(5) \textbf{\emph{Foundation model-based approaches}}: UniST~\cite{ref:UniST}, which pretrains a general spatio-temporal model on diverse datasets, and ST-LLM~\cite{liu2024spatial}, which adapts large language models for ST forecasting.

\paragraph{Implementation Details}
Following prior work~\cite{ref:UniST}, we set both the historical window and prediction horizon to 6 time steps, requiring the model to forecast the next 6 steps given the past 6 observations. To ensure consistency across baselines, we uniformly patch all spatial and temporal inputs using a patch size of 2 and apply the same processing to the ground-truth outputs.

The number of cross-prompt attention layers in the Bilateral Cross-Prompt Coordination module is set to 2 for most datasets, except for BikeNYC and TaxiNYC where it is reduced to 1 due to smaller data scale. The Temporal Transformer blocks in both the Pre-VFM and Post-VFM Reprogramming stages are set to 3 layers.

We adopt DINO~\cite{dino2} as the primary VFM backbone and use the AdamW optimizer, selecting the best checkpoint based on validation performance. The VFM backbone remains frozen during training, while the Pre-VFM and Post-VFM Reprogramming modules are trained end-to-end. Additionally, a lightweight Adapter~\cite{Adapter} is inserted into the VFM backbone to allow optional parameter-efficient fine-tuning, though it is not activated in all configurations.

All experiments are conducted using PyTorch on an NVIDIA RTX A5500 GPU with 24 GB memory.

\begin{table*}[t]
\centering
\setlength{\tabcolsep}{1.6pt}
\resizebox{\textwidth}{!}{%
\begin{tabular}{l|l|l|l|cc|cc|cc|cc|cc}
\hline
\multirow{2}{*}{Method/ID} & 
\multirow{2}{*}{Encoder} & 
\multirow{2}{*}{Pre-VFM Reprogram} & 
\multirow{2}{*}{Post-VFM Reprogram} & 
\multicolumn{2}{c|}{Crowd} & 
\multicolumn{2}{c|}{Cellular} & 
\multicolumn{2}{c|}{TDrive} & 
\multicolumn{2}{c|}{BikeNYC} & 
\multicolumn{2}{c}{TaxiNYC} \\ \cline{5-14}
& & & & RMSE & MAE & RMSE & MAE & RMSE & MAE & RMSE & MAE & RMSE & MAE \\ \hline

\textbf{\#1 (simple)} & \emph{VFM} & \emph{Simple Adaptation} & \emph{MLP} & 10.51 & 4.14 & 43.11 & 16.93 & 94.73 & 47.24 & 8.49 & 2.65 & 39.18 & 13.20 \\
\textbf{UniST} & \emph{Transformer} & \emph{Knowledge-guided Prompt} & \emph{Transformer} & 3.00 & 1.38 & 14.29 & 6.50 & 44.97 & 19.76 & 3.50 & 1.27 & 15.32 & 5.65 \\
\textbf{\#2} & \emph{VFM} & \emph{Token Adaptation} & \emph{Transformer} & 2.69 & 1.22 & 13.37 & 5.93 & 42.82 & 18.25 & 3.53 & 1.29 & 14.69 & 5.62 \\
\textbf{\#3} & \emph{VFM} & \emph{+Temporal Context Encoder} & \emph{Transformer} & 2.63 & 1.17 & 12.95 & 5.78 & 40.65 & 17.36 & 3.44 & 1.25 & 14.38 & 5.41 \\
\textbf{\#4} & \emph{VFM+Adapter} & \emph{+Temporal Context Encoder} & \emph{Transformer} & 2.65 & 1.17 & 12.84 & 5.65 & 40.02 & 17.21 & \textbf{3.36} & 1.22 & 13.98 & 5.31 \\
\textbf{Ours (full)} & \emph{VFM+Adapter} & \emph{+Temporal Context Encoder+Flow} & \emph{+Cross-Prompt Coordination} & \textbf{2.51} & \textbf{1.06} & \textbf{12.31} & \textbf{5.57} & \textbf{39.54} & \textbf{16.94} & 3.38 & \textbf{1.20} & \textbf{13.79} & \textbf{4.95} \\ \hline
\end{tabular}
}
\vspace{-0.6em}
\caption{Ablation study on model components. In \textbf{\#1 (simple)}, simply adopting VFM fails. Comparing \textbf{UniST} and \textbf{\#2}, the VFM encoder outperforms the UniST pretrained Transformer. With our novel Pre-VFM and Post-VFM reprogramming designs, \textbf{Ours (full)} shows substantial improvements.}
\label{tab:ablation}
\vspace{-1.2em}
\end{table*}

\subsection{Main Results}

The results in Table~\ref{tab:result} yield several key observations: \\
(1)~\textbf{Conventional statistical models} (HA, ARIMA) consistently deliver the weakest performance across all datasets. Their inability to model nonlinear patterns or capture spatial dependencies makes them inadequate for complex spatio-temporal forecasting tasks. \\ 
(2)~\textbf{Multivariate time series forecasting models} (PatchTST, iTransformer) perform notably worse than spatio-temporal models, confirming that ignoring spatial structures—even when using advanced temporal architectures—leads to suboptimal predictions. This highlights the critical role of integrating spatial information alongside temporal modeling. \\
(3)~\textbf{Video prediction models} (PredRNN, MAU, MIM, SimVP, TAU) achieve performance close to customized spatio-temporal forecasting models (e.g., STID), reflecting the shared nature of these tasks: both require capturing evolving spatio-temporal dynamics. Their results further support the rationale for leveraging VFMs, as both video and spatio-temporal forecasting benefit from strong visual representations. \\
(4)~\textbf{Foundation model-based approaches}, including UniST and ST-LLM, show competitive performance, with UniST standing out as the strongest prior spatio-temporal baseline. UniST relies on large-scale pretraining across spatio-temporal datasets, followed by dataset-specific fine-tuning. In contrast, our approach reprograms VFMs pretrained solely on large-scale image data, eliminating the need for costly spatio-temporal pretraining, yet still outperforming UniST and ST-LLM. This emphasizes the advantage of transferring general visual knowledge to spatio-temporal tasks. \\
(5)~\textbf{Our ST-VFM consistently achieves the best results across all datasets and metrics}, outperforming all baselines on both MAE and RMSE. Notably, ST-VFM improves over the best baseline by more than $10\%$ on average. This demonstrates the effectiveness of our approach: combining the spatial priors of image-pretrained VFMs with explicit temporal modeling using lightweight motion proxies leads to superior forecasting accuracy across diverse domains.  

\subsection{Ablation Studies}

\paragraph{Effectiveness of each component}  
We evaluate the contribution of each component in ST-VFM through an ablation study (Table~\ref{tab:ablation}). 
We begin with \textbf{\#1 (Simple)}, which combines a vision foundation model (DINO) with an MLP decoder. This variant yields the weakest performance, showing that VFMs alone, without temporal modeling, are insufficient for spatio-temporal forecasting.

We then compare to \textbf{UniST}, a strong baseline that employs a Transformer encoder-decoder pretrained on spatio-temporal data. 
Replacing UniST’s encoder with our VFM plus a basic token adaptation layer (\textbf{\#2}) already improves results, highlighting the strong spatial representation capacity of VFMs with our tailored adaptation.

Next, adding a Temporal Context Encoder in the pre-reprogramming stage (\textbf{\#3}) further improves performance, confirming the importance of modeling temporal dependencies before spatial encoding. Incorporating lightweight adapters into the VFM backbone (\textbf{\#4}), inserted between the self-attention and residual connections, provides additional fine-tuning flexibility and leads to further gains. 

Finally, introducing the ST Flow input alongside the Bilateral Cross-Prompt Coordination module (\textbf{Ours (Full)}) achieves the best overall performance. This shows that integrating motion-aware auxiliary signals and dynamic cross-branch interaction enables VFMs to effectively capture spatio-temporal correlations, extending their capacity beyond static spatial modeling.

\begin{figure}[t]
    \centering
    \includegraphics[width=0.85\linewidth]{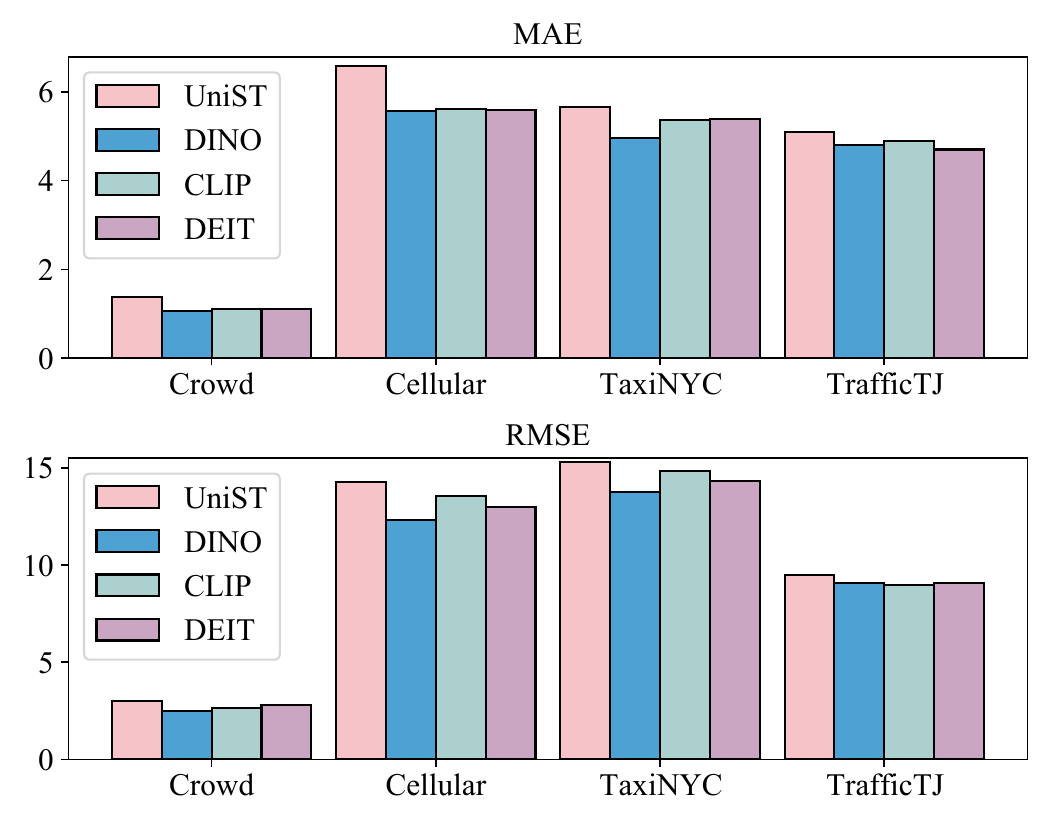}
    \caption{Comparison of different pretrained VFMs (DINO, CLIP, DEIT) on representative spatio-temporal datasets.\protect\footnotemark}
    \label{fig:VFM_Variants}
\end{figure}
\footnotetext{To plot all datasets on the same scale, we multiply the values of the TrafficTJ dataset by a factor of 10.}
\paragraph{Performance with Different VFMs}
To evaluate the generality of our approach across different vision foundation models, we test three representative backbones pretrained with distinct paradigms: \emph{self-supervised} learning (DINO~\cite{dino2}), \emph{contrastive learning} (CLIP~\cite{ref:CLIP}), and \emph{supervised learning} (DEIT~\cite{ref:DEIT}). As shown in Figure~\ref{fig:VFM_Variants}, all three VFM variants consistently outperform the strongest prior baseline, UniST, across four representative datasets. Minor differences in performance across datasets can be attributed to differences in the underlying pretraining strategies. Detailed descriptions of the VFM architectures and comprehensive results on all datasets are provided in the Appendix.

\paragraph{Video Foundation Models}
One might wonder why not directly use video foundation models (Video FMs) to handle spatial and temporal correlations, given their pretraining on large-scale video data. We explore this by adapting VideoMAE~\cite{ref:videoMAE}, a representative Video FM, using a similar adaptation pipeline as for image-based VFMs. As shown in Table~\ref{tab:video}, VideoMAE performs better than conventional video prediction models (e.g., SimVP), confirming the benefit of pretrained visual representations. 
However, it still underperforms compared to our reprogrammed VFMs. This performance gap likely arises from two factors: (1) the relatively smaller scale and lower diversity of video pretraining datasets compared to image datasets, which limits the richness of learned spatial priors; and (2) the inherent entanglement of spatial and temporal representations in Video FMs, which reduces flexibility in disentangling motion patterns critical for forecasting tasks.

\begin{table}[t]
    \centering
    \setlength{\tabcolsep}{1.5pt}
    \resizebox{0.5\textwidth}{!}{
    \begin{tabular}{c|cccccccccc}
    \hline
         \multirow{2}{*}{Model} & \multicolumn{2}{c}{Crowd} & \multicolumn{2}{c}{Cellular} & \multicolumn{2}{c}{TDrive} & \multicolumn{2}{c}{BikeNYC} & \multicolumn{2}{c}{TaxiNYC}  \\ \cline{2-11}
         & RMSE & MAE & RMSE & MAE & RMSE & MAE & RMSE & MAE & RMSE & MAE \\ \hline
         SimVP& 3.91 & 1.96 & 16.48 & 8.23 & 46.80 & 22.90 & 4.11 & 1.67 & 16.63 & 7.51 \\\hline
         Video& 2.90 & 1.27 & 15.58 & 6.08 & 41.09 & 17.78 & 3.91 & 1.42 & 16.64 & 6.00 \\ \hline
        VFM&  \textbf{2.51}& \textbf{1.06} & \textbf{12.31} & \textbf{5.57} & \textbf{39.54} & \textbf{16.94} & \textbf{3.38} & \textbf{1.20} & \textbf{13.79} & \textbf{4.95} \\ \hline
    \end{tabular}
    }
\caption{\label{tab:video}Comparison between image-based VFMs and the video foundation model VideoMAE on representative spatio-temporal datasets. While VideoMAE surpasses conventional video prediction models (e.g., SimVP), it is consistently outperformed by the reprogrammed image-based VFMs, highlighting the stronger spatial priors and flexibility of image-pretrained models for spatio-temporal forecasting.}
\end{table}

\section{Conclusion}

In this paper, we propose ST-VFM, a novel framework that reprograms vision foundation models for spatio-temporal forecasting. ST-VFM leverages the strong spatial priors of large-scale pretrained VFMs and extends them with lightweight temporal modeling components, including a temporal context encoder, ST flow input as a motion proxy, and a bilateral cross-prompt coordination module for dynamic spatial-temporal interaction. Additionally, an auxiliary flow forecasting task is introduced to enhance temporal representation learning and regularize the main forecasting objective. Through extensive experiments on ten diverse datasets, ST-VFM consistently outperforms state-of-the-art baselines across multiple domains and tasks. Our results demonstrate that with carefully designed reprogramming, pretrained VFMs can be effectively adapted to spatio-temporal tasks without requiring costly task-specific pretraining, offering a promising direction for future research on general-purpose foundation models in temporal domains.

\bibliography{aaai2026}

\end{document}


\maketitle

\section{Loss Function}
The overall loss of ST-VFM combines the Mean Squared Error (MSE) from two forecasting tasks: the spatio-temporal (ST) prediction and the auxiliary flow prediction. A balancing weight $\lambda$ controls the contribution of the flow loss:
\begin{align}
\mathcal{L}_{\mathrm{ST}} &= \frac{1}{THW} \sum_{t=1}^{T} \sum_{h=1}^{H} \sum_{w=1}^{W} \| \hat{Y}^{\mathrm{ST}}_{thw} - Y^{\mathrm{ST}}_{thw} \|^2 \,, \\
\mathcal{L}_{\mathrm{Flow}} &= \frac{1}{THW} \sum_{t=1}^{T} \sum_{h=1}^{H} \sum_{w=1}^{W} \| \hat{Y}^{\mathrm{Flow}}_{thw} - Y^{\mathrm{Flow}}_{thw} \|^2 \,, \\
\mathcal{L} &= \mathcal{L}_{\mathrm{ST}} + \lambda \mathcal{L}_{\mathrm{Flow}} \,.
\end{align}

\section{Evaluation Metrics}
We evaluate forecasting performance using Mean Absolute Error (MAE) and Root Mean Squared Error (RMSE). Given ground truth values $Y_1, \dots, Y_N$ and predicted values $\hat{Y}_1, \dots, \hat{Y}_N$, where $N$ is the total number of spatio-temporal points, the metrics are computed as:
\begin{align}
    \mathrm{MAE}(Y, \hat{Y}) &= \frac{1}{N} \sum_{i=1}^N |Y_i - \hat{Y}_i| \,, \\
    \mathrm{RMSE}(Y, \hat{Y}) &= \sqrt{\frac{1}{N} \sum_{i=1}^N (Y_i - \hat{Y}_i)^2} \,.
\end{align}

\section{More Experimental Results}

\begin{figure}[t]
\centering
\subfigure[Cellular]{\label{1}\includegraphics[width=0.22\textwidth]{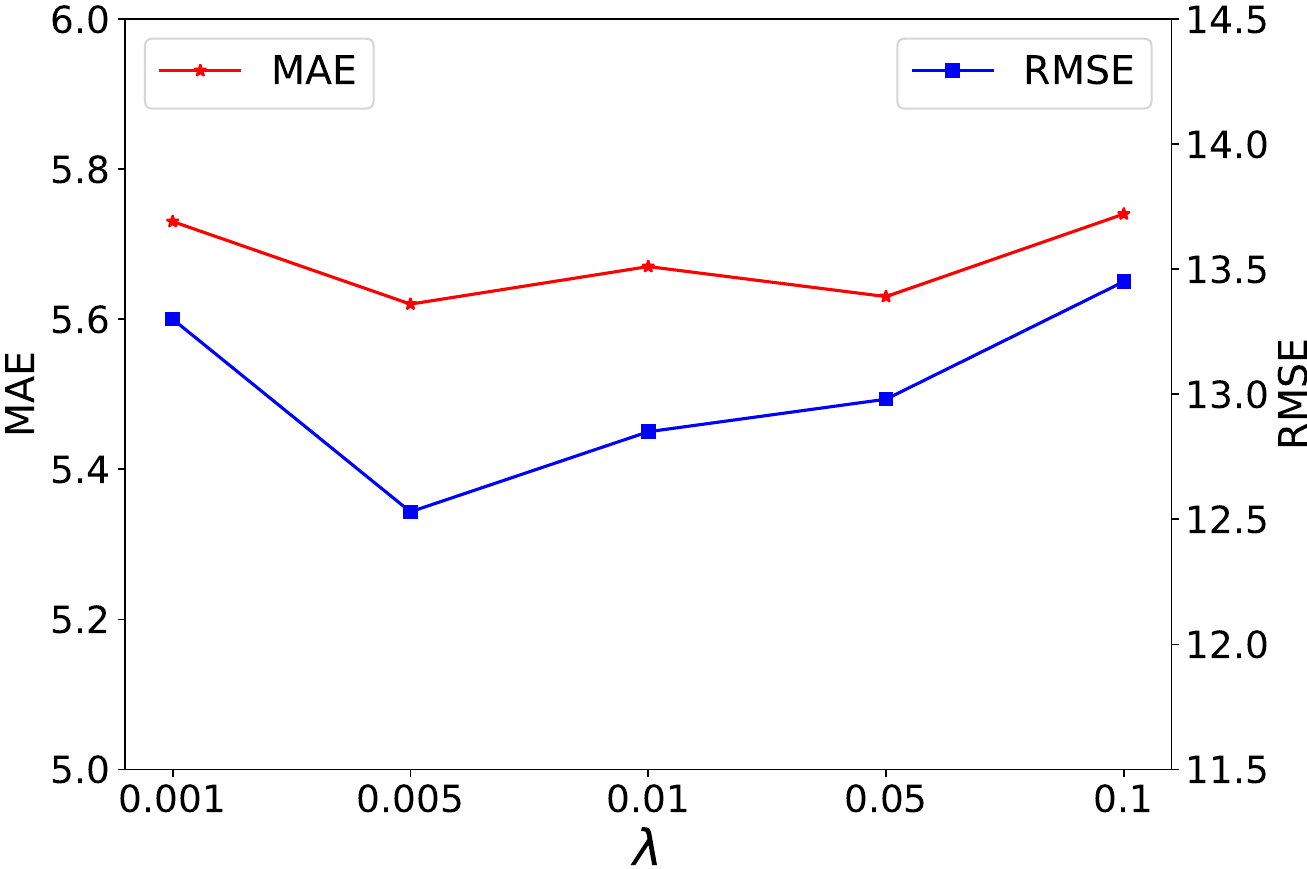}} 
\subfigure[BikeNYC]{\label{2}\includegraphics[width=0.22\textwidth]{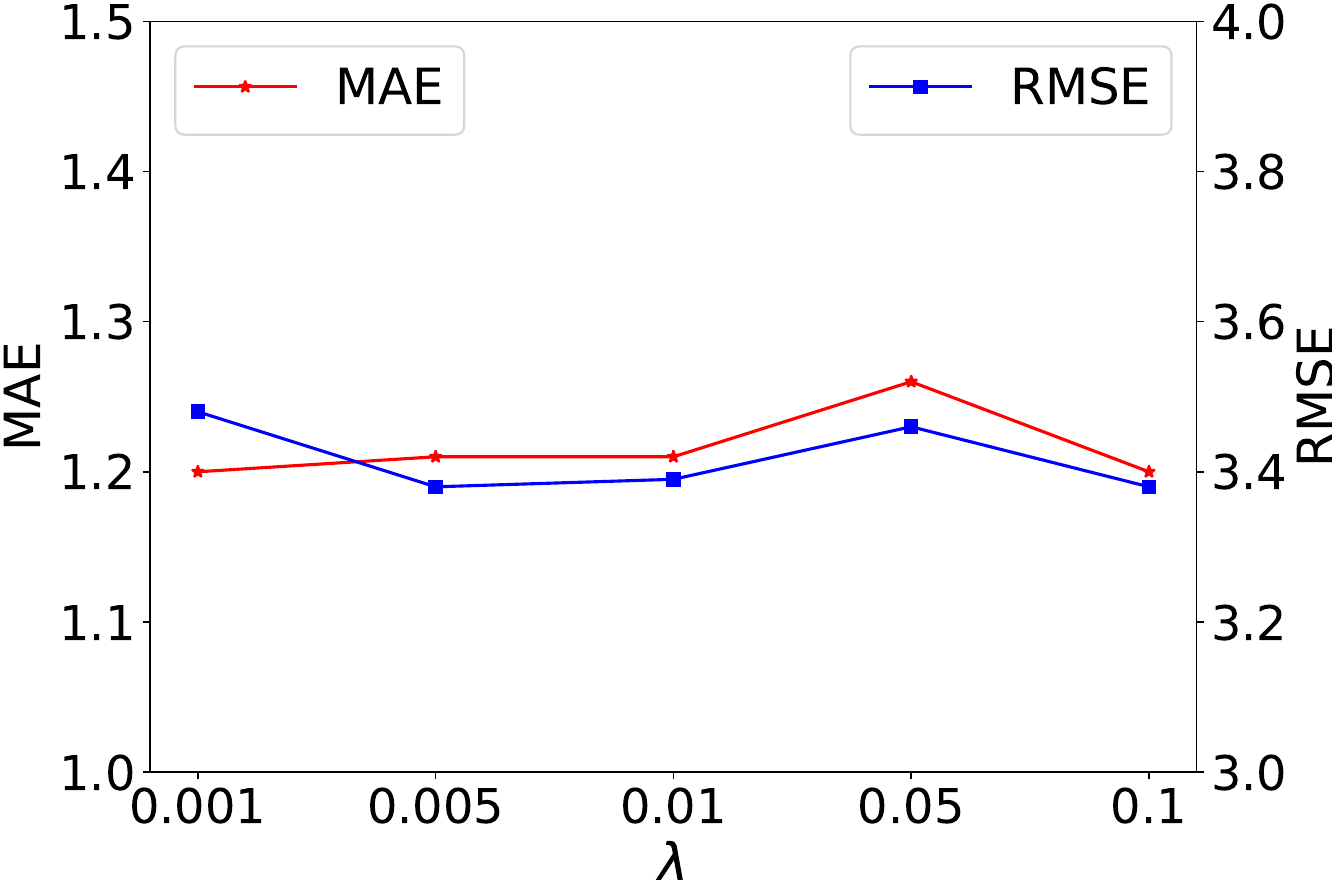}}  
\caption{Sensitivity  w.r.t. $\lambda$ on Cellular and BikeNYC.} 
\label{fig:para}
\end{figure}

\paragraph{Parameter Analysis}
We analyze the impact of the loss weight $\lambda$ on two representative datasets. As shown in Figure~\ref{fig:para}, our method maintains stable performance across a wide range of $\lambda$ values on both Cellular and BikeNYC, demonstrating robustness and adaptability to this hyperparameter.

\paragraph{Efficiency-Performance Trade-Off}
We further analyze the trade-off between efficiency and performance. As shown in Table~\ref{tab:time}, our model variant (\textbf{\#4}) achieves significantly better forecasting performance than the state-of-the-art UniST while maintaining comparable training and inference times. The full ST-VFM model further improves performance with only a moderate increase in computational cost, demonstrating an effective balance between accuracy and efficiency. 
All experiments are conducted with Pytorch on an NVIDIA RTX A5500 GPU with 24G memory.

\begin{table}[t]
\centering
\resizebox{0.45\textwidth}{!}{%
\begin{tabular}{c|cccc}
\hline
& Training (min)& Inference (min)  & RMSE  &MAE \\ \hline
\textbf{UniST} &  1.06 & 0.0423 & 3.00 & 1.38\\ \hline
\textbf{\# 4}  &  1.08 & 0.0432 & 2.65 & 1.17 \\ \hline
\textbf{Ours (full)} &  2.06 & 0.0745 & 2.51 & 1.06\\ \hline
\end{tabular}
}
\vspace{-0.5em}
\caption{Efficiency Comparison on Crowd dataset.}
\label{tab:time}
\vspace{-1.2em}
\end{table}

\begin{table*}[t]
    \centering
    \resizebox{1.0\textwidth}{!}{
    \begin{tabular}{c|cccccccccccccccccccc}
    \hline
         \multirow{2}{*}{Model} & \multicolumn{2}{c}{Crowd} & \multicolumn{2}{c}{Cellular} & \multicolumn{2}{c}{TDrive} & \multicolumn{2}{c}{BikeNYC} & \multicolumn{2}{c}{TaxiNYC} & \multicolumn{2}{c}{BikeNYC2} & \multicolumn{2}{c}{TaxiNYC2} & \multicolumn{2}{c}{TrafficZZ} & \multicolumn{2}{c}{TrafficTJ} & \multicolumn{2}{c}{TrafficSH}  \\ \cline{2-21}
         & RMSE & MAE & RMSE & MAE & RMSE & MAE & RMSE & MAE & RMSE & MAE & RMSE & MAE & RMSE & MAE & RMSE & MAE & RMSE & MAE & RMSE & MAE \\ \hline
        SimVP & 3.91 & 1.96 & 16.48 & 8.23 & 46.80 & 22.90 & 4.11 & 1.67 & 16.63 & 7.51 &5.96 &2.92 &15.10 &6.54 &0.838 &0.526 &1.00 &0.597 &0.814 & 0.569 \\ \hline
        Video & 2.90 & 1.27 & 15.58 & 6.08 & 41.09 & 17.78 & 3.91 & 1.42 & 16.64 & 6.00 &5.71 &2.62 &15.25 &5.13 &0.815 &0.458 &0.937 &0.492 &0.747 & 0.486 \\ \hline
         VFM & \textbf{2.51} & \textbf{1.06} & \textbf{12.31} & \textbf{5.57} & \textbf{39.54} & \textbf{16.94} & \textbf{3.38} & \textbf{1.20} & \textbf{13.79} & \textbf{4.95} & \textbf{5.42} & \textbf{2.46} & \textbf{11.85} & \textbf{4.39} & \textbf{0.800} & \textbf{0.457} & \textbf{0.909} & \textbf{0.484} & \textbf{0.645} & \textbf{0.392} \\ \hline
    \end{tabular}
    }
    
\caption{Comparison between image-based VFMs and the video foundation model (VideoMAE) on all spatio-temporal datasets; VideoMAE also outperforms conventional video prediction models (e.g., SimVP).}
    \label{tab:video_1}
\end{table*}

\begin{table*}[t]
    \centering
    \resizebox{1.0\textwidth}{!}{
    \begin{tabular}{c|cccccccccccccccccccc}
    \hline
         \multirow{2}{*}{Model} & \multicolumn{2}{c}{Crowd} & \multicolumn{2}{c}{Cellular} & \multicolumn{2}{c}{TaxiNYC} & \multicolumn{2}{c}{TrafficTJ} & \multicolumn{2}{c}{TDrive} & \multicolumn{2}{c}{BikeNYC} & \multicolumn{2}{c}{BikeNYC2} & \multicolumn{2}{c}{TaxiNYC2} & \multicolumn{2}{c}{TrafficZZ} & \multicolumn{2}{c}{TrafficSH}  \\ \cline{2-21}
         & RMSE & MAE & RMSE & MAE & RMSE & MAE & RMSE & MAE & RMSE & MAE & RMSE & MAE & RMSE & MAE & RMSE & MAE & RMSE & MAE & RMSE & MAE \\ \hline
         UniST  &3.00 &1.38 &14.29 &6.59 &15.32 &5.65 &0.95 &0.51 &44.97 & 19.67 & 3.50 & 1.27 &5.50 &2.56 &12.71 &4.82 & 0.832 & 0.482 &0.665 &0.405 \\ \hline
         CLIP  &2.66 &1.10 &13.57 &5.61 &14.84 &5.36 &\textbf{0.90} &0.49 &N/A &N/A &3.45 &1.20 &5.42 &2.58 &11.77  &\textbf{4.19}  &0.814 &0.469 &0.644 &0.396  \\ \hline
         DEIT  &2.78 &1.10 &12.99 &5.60 &14.35 &5.38 &0.91 &\textbf{0.47} &N/A &N/A &3.49 &1.21 &\textbf{5.34} &2.55  &11.90 & 4.22  &0.821 &0.467 &\textbf{0.641} & \textbf{0.387}  \\ \hline
         DINO  &\textbf{2.51} &\textbf{1.06} &\textbf{12.31} &\textbf{5.57} &\textbf{13.79} &\textbf{4.95} &0.91 &0.48 &\textbf{39.54} &\textbf{16.94} &\textbf{3.38} &\textbf{1.20} &5.42 &\textbf{2.46} &\textbf{11.85} &4.39 & \textbf{0.800} & \textbf{0.457} & 0.645& 0.392  \\ \hline
    \end{tabular}
    }
    \caption{Comparison of pretrained VFMs (DINO, CLIP, DEIT) with UniST on all spatio-temporal datasets.}
    \label{tab:different_VFM}
\end{table*}

\paragraph{Complete Results of Video and Vision Foundation Models}
We provide the full results of the video foundation model and various vision foundation models across all datasets in Table~\ref{tab:video_1} and Table~\ref{tab:different_VFM}. These extended results support the same conclusions reported in the main paper, further validating the robustness and consistency of our findings.

\section{Baselines and State-of-the-Art Methods}
We compare our model with a broad range of baselines spanning statistical models, spatio-temporal deep learning methods, video prediction approaches, and recent advances in pretraining and meta-learning.

\begin{itemize}
  \item \textbf{HA (Historical Average)}: Predicts future values using the average of historical data from corresponding periods in previous days. This serves as a simple yet informative non-learning baseline.

  \item \textbf{ARIMA}: A classical statistical model for time series forecasting, which captures auto-regressive, integrated, and moving average components. It is widely used for its interpretability and effectiveness in modeling linear temporal dynamics.

  \item \textbf{STResNet}~\cite{zhang2017deep}: A spatio-temporal residual network designed for crowd flow prediction. It models temporal closeness, periodicity, and long-term trends using separate residual structures.

  \item \textbf{ACFM}~\cite{liu2018attentive}: The Attentive Crowd Flow Machine introduces an attention mechanism to adaptively aggregate sequential and periodic patterns for more accurate flow prediction.

  \item \textbf{STGSP}~\cite{zhao2022st}: Emphasizes the importance of global and positional information in the temporal dimension. It employs a semantic flow encoder and multi-scale temporal attention to improve forecasting performance.

  \item \textbf{MC-STL}~\cite{zhang2023mask}: Utilizes masked contrastive learning for spatio-temporal representation learning, which enhances the model’s ability to capture relationships across space and time.

  \item \textbf{MAU}~\cite{chang2021mau}: A motion-aware unit designed for video prediction. It expands temporal receptive fields to better capture inter-frame motion dynamics using a combination of attention and fusion modules.

  \item \textbf{PredRNN}~\cite{wang2017predrnn}: A recurrent network with explicitly decoupled memory cells and zigzag memory flow. This architecture helps model spatio-temporal dependencies at multiple levels.

  \item \textbf{MIM}~\cite{wang2019memory}: The Memory In Memory network captures non-stationary dynamics in sequences by modeling differential information between recurrent states, with stacked blocks for higher-order dynamics.

  \item \textbf{SimVP}~\cite{gao2022simvp}: A simple yet effective video prediction model based entirely on convolutional architectures. It achieves strong performance despite its minimal design.

  \item \textbf{TAU}~\cite{tan2023temporal}: The Temporal Attention Unit decomposes attention into intra-frame (static) and inter-frame (dynamic) components. It introduces differential divergence regularization to handle inter-frame variations.

  \item \textbf{STID}~\cite{shao2022spatial}: A lightweight MLP-based spatio-temporal model. It highlights the challenge of indistinguishability among spatio-temporal samples and demonstrates that well-designed MLPs can be highly competitive.

  \item \textbf{STNorm}~\cite{deng2021st}: Proposes two specialized normalization modules—spatial and temporal normalization—to separately handle local and high-frequency components in the data.

  \item \textbf{PatchTST}~\cite{nie2022time}: Segments time series into patches to capture long-term dependencies and processes each channel independently. It also leverages self-supervised learning for enhanced generalization.

  \item \textbf{iTransformer}~\cite{liu2023itransformer}: A Transformer-based model that applies attention mechanisms on the inverted (channel) dimension, effectively modeling inter-variable correlations in multivariate time series.

  \item \textbf{ST-LLM}~\cite{liu2024spatial}: A method which treats each graph node as a token from the graph’s perspective, enabling the model to represent semantic information for both spatial and temporal correlations.

  \item \textbf{UniST}~\cite{ref:UniST}: A unified framework for spatio-temporal forecasting that reformulates various prediction tasks as masked sequence modeling problems. UniST introduces a prompt-based tuning strategy to adapt a shared backbone across different spatio-temporal tasks and datasets, leveraging pre-training and flexible prompting to enhance generalization and transferability.
\end{itemize}

\section{Vision Foundation Models}
In our experiments, we leverage several state-of-the-art Vision Foundation Models (VFMs) as backbone architectures for spatial representation learning. These models are selected for their diverse pretraining paradigms and demonstrated generalization across downstream visual tasks.

\paragraph{DINO (Self-Distillation with No Labels)}
DINO~\cite{dino,dino2} is a self-supervised learning framework based on Vision Transformers (ViTs). It employs a student–teacher setup optimized with a self-distillation loss, where both networks process distinct augmentations of the same image and are trained to produce similar representations. DINO effectively learns semantically meaningful and linearly separable features without requiring labeled data. We adopt DINOv2 with the small ViT backbone.

\paragraph{CLIP (Contrastive Language–Image Pretraining)}
CLIP~\cite{ref:CLIP} is a vision–language model trained to align visual and textual representations in a shared embedding space using contrastive learning on large-scale web data. While primarily developed for zero-shot image classification, its visual encoder (typically ViT or ResNet) provides rich, transferable semantic features for vision-only tasks. We use the base ViT model as the visual encoder.

\paragraph{DEIT (Data-Efficient Image Transformer)}
DEIT~\cite{ref:DEIT} is a supervised ViT variant designed for improved data efficiency. It introduces a distillation token, enabling the model to leverage a convolutional teacher (e.g., RegNet) during training. Despite being trained only on ImageNet-1K, DEIT achieves strong performance, making it an efficient and competitive baseline. We use the base ViT model as the visual encoder.

\paragraph{VideoMAE (Masked Autoencoding for Video)}
VideoMAE~\cite{ref:videoMAE} extends the masked autoencoder paradigm to video data. It performs spatiotemporal masked reconstruction using a ViT encoder and a lightweight decoder, applying cube masking strategies to jointly model spatial and temporal dynamics. VideoMAE’s pretraining produces robust representations transferable to a range of video understanding tasks. We adopt VideoMAEv1 with the small ViT backbone.

\bibliography{aaai2026}